\documentclass[conference]{IEEEtran}

\usepackage{cite}
\usepackage{amsmath,amssymb,amsfonts}

\usepackage{graphicx}
\usepackage{textcomp}
\usepackage{xcolor}
\def\BibTeX{{\rm B\kern-.05em{\sc i\kern-.025em b}\kern-.08em
    T\kern-.1667em\lower.7ex\hbox{E}\kern-.125emX}}

\usepackage{cite}
\usepackage[utf8]{inputenc}
\usepackage[T1]{fontenc}
\usepackage{amsmath,amssymb,amsfonts}

\usepackage{graphicx}
\usepackage{textcomp}

\UseRawInputEncoding
\PassOptionsToPackage{bookmarks=false, hidelinks}{hyperref}
\usepackage{hyperref}
\usepackage{cleveref}
\usepackage{booktabs}
\usepackage{comment}
\usepackage{tabularx}
\usepackage{nicematrix}
\usepackage{adjustbox}
\usepackage{afterpage}

\usepackage{algorithm}
\usepackage{algpseudocode}

\usepackage[english]{babel}
\def\BibTeX{{\rm B\kern-.05em{\sc i\kern-.025em b}\kern-.08em
    T\kern-.1667em\lower.7ex\hbox{E}\kern-.125emX}}

\addto\extrasenglish{%

}

\begin{document}
\UseRawInputEncoding
\title{
\LARGE \textbf{Point Cloud Recombination: Systematic Real Data Augmentation Using Robotic Targets for LiDAR Perception Validation}\\
\vspace{1 mm}
}

\author{
    Hubert Padusinski$^{1}$, Christian Steinhauser$^{1}$, Christian Scherl$^{1}$, Julian Gaal$^{2}$, Jacob Langner$^{1}$ \\
    \small $^1$FZI Research Center for Information Technology, Karlsruhe, Germany \\
    \small $^2$ANavS GmbH - Advanced Navigation Solutions, M{\"u}nchen, Germany \\
    \small \{padusinski, steinhauser, scherl, langner\}@fzi.de, julian.gaal@anavs.de
}

\thanks{$^{1}$Hubert Padusinski, Christian Steinhauser, Christian Scherl and Jacob Langner are with FZI Research Center for Information Technology, 76131 Karlsruhe, Germany. 
        \{\tt\small padusinski, steinhauser, scherl, langner\}@fzi.de} \thanks{$^{2}$Julian Gaal is with ANavS GmbH - Advanced Navigation Solutions, 80686 M{\"u}nchen, Germany.
         {\tt\small julian.gaal@anavs.de}}

%\bstctlcite{IEEEexample:BSTcontrol}
\maketitle
\thispagestyle{empty}
\pagestyle{empty}

%%%%%%%%%%%%%%%%%%%%%%%%%%%%%%%%%%%%%%%%%%%%%%%%%%%%%%%%%%%%%%%%%%%%%%%%%%%%%%%%
\begin{abstract}
The validation of LiDAR-based perception of intelligent mobile systems operating in open-world applications remains a challenge due to the variability of real environmental conditions. Virtual simulations allow the generation of arbitrary scenes under controlled conditions but lack physical sensor characteristics, such as intensity responses or material-dependent effects. In contrast, real-world data offers true sensor realism but provides less control over influencing factors, hindering sufficient validation. Existing approaches address this problem with augmentation of real-world point cloud data by transferring objects between scenes. However, these methods do not consider validation and remain limited in controllability because they rely on empirical data. We solve these limitations by proposing Point Cloud Recombination, which systematically augments captured point cloud scenes by integrating point clouds acquired from physical target objects measured in controlled laboratory environments. Thus enabling the creation of vast amounts and varieties of repeatable, physically accurate test scenes with respect to phenomena-aware occlusions with registered 3D meshes. Using the Ouster OS1-128 Rev7 sensor, we demonstrate the augmentation of real-world urban and rural scenes with humanoid targets featuring varied clothing and poses, for repeatable positioning. We show that the recombined scenes closely match real sensor outputs, enabling targeted testing, scalable failure analysis, and improved system safety. By providing controlled yet sensor-realistic data, our method enables trustworthy conclusions about the limitations of specific sensors in compound with their algorithms, e.g., object detection.

\end{abstract}
%------------------------------------------------------------%
\section{Introduction}
\label{sec:introduction}
LiDAR-based perception enables intelligent mobile systems from cars ~\cite{lidarsurveyauto, heinrich2024cocarnextgenmultipurposeplatform} and buses, construction and agricultural machinery ~\cite{Petereit2019, agrirobot}. They offer spatial and material-sensitive information in the functional cause chain, or are used as reference systems~\cite{fernandes2021realtime} for ground truth generation. Validating LiDAR-based functions requires test data that reflect both representative and critical operating conditions across defined Operational Design Domains (ODDs)~\cite{jan_2016_CVPR, Heide2020b, mviceberg, kuznietsov2025methodologystatisticalanalysisinfluencing}. Explorative real world data acquisition \cite{kitti, waymodataset} provides realistic LiDAR point cloud data (PCD) under authentic settings. However, the combinatorial dimensionality of high-resolution data necessitates sparse sampling on how scenes can appear, limiting variability and comparability in collected datasets. Simulations, by contrast, allow controlled variation but often fail to replicate sensor-specific physical effects, such as intensity values or local reflectivity phenomena. 
\begin{figure}[ht]
    \centering
    %\vspace{-11mm}
    \includegraphics[width=1.0\linewidth]{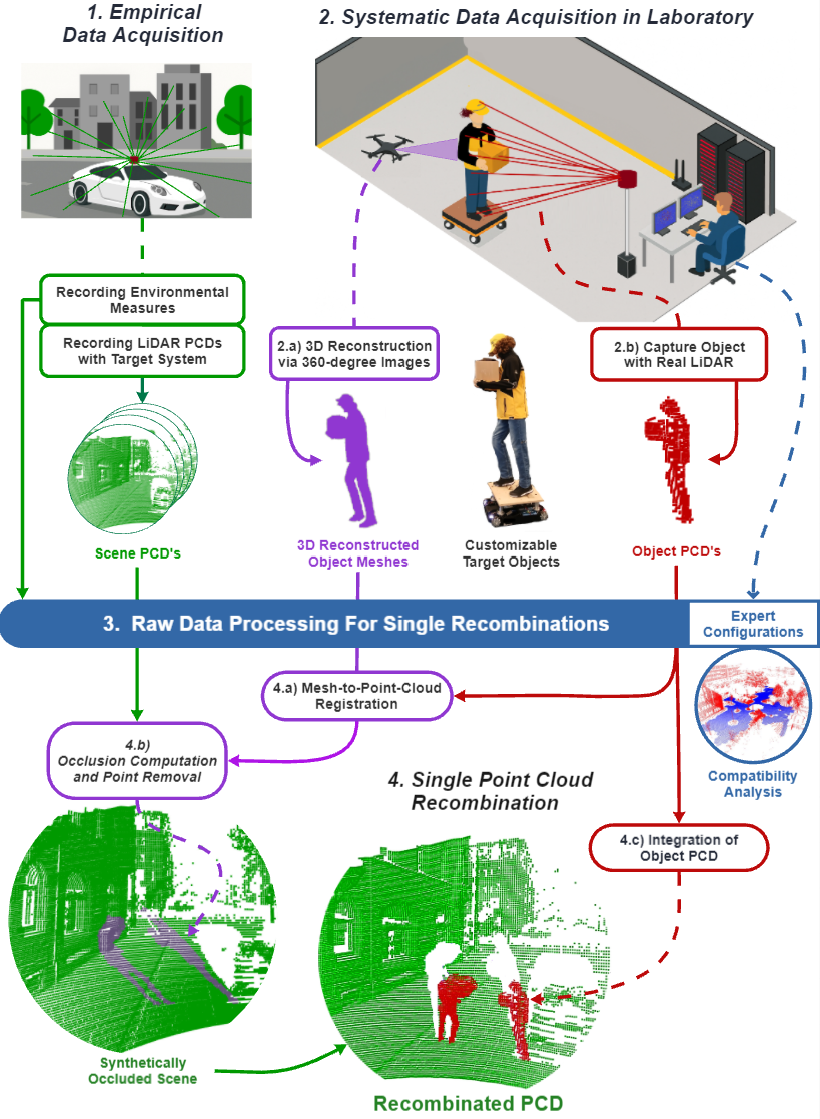}
    \vspace{-1mm}
    \caption{PCR method for modifying PCD scenes from the ODD using PCD objects and 3D meshes systematically recorded from target objects in the laboratory.}
    \label{fig:pcr_intro}
    \vspace{-7mm}
\end{figure}

\textit{Point Cloud Recombination (PCR)} bridges virtual and real drive test, as a hardware perception simulation method that enables controlled augmentation of real scenes using real data from physical target objects, thereby enabling diverse and validation appropriate test data. Combined with automation solutions such as mobile robots, PCR facilitates scalable and repeatable data acquisition from the sensor under test with variable positioning. 3D meshes of physical targets are aligned with the corresponding point clouds to enable plausible phenomena-aware occlusion computation via raycasting, especially when real object data exhibit artifacts.The method was evaluated using an Ouster OS1-128 Rev7 LiDAR and humanoid targets mounted on a mobile platform, with data acquired at relative distances of up to 35 m.

\section{Motivation behind Systematic Augmentation}
\label{sec:motivation}

Studies have shown that certain isolated factors within the ODD, particularly the characteristic
of a whole scene ~\cite{jan_2016_CVPR} or specific object attributes ~\cite{kuznietsov2025methodologystatisticalanalysisinfluencing}, affect LiDAR functions. However, observing
factors in isolation is insufficient to explain the underlying causes of failures. Research in Explainable AI further highlights the need to consider the logical compositions of factors, as objects and their properties cannot be observed in isolation from their surrounding context
~\cite{xiao2021noise, Oksuz2021}. In our past work \cite{mviceberg}, we propose to examine ODDs at multiple granularities. We define a
scene $\mathcal{S}$ consisting of granularity orders $\mathcal{GO}_1, \dots, \mathcal{GO}_n$, with
each order $\mathcal{GO}_n$ containing entities $e \in \mathcal{GO}_n$. Each entity is associated with
discrete or continuous properties $\phi_i(e)$ and may relate to super-- or sub--ordinate entities. 
Varying single entities or properties, under otherwise constant plausible conditions (\textit{ceteris-paribus}), allows us to analyze combinations of factors that lead to failures. Under this assumption we define systematic augmentation of a scene $\mathcal{S}^{+\delta}$ results from applying a modification $\delta$ via four operations:  
\vspace{-4mm}
\begin{align}
\delta_{\text{modification}} &:= \left( \phi_i(e) \mapsto \phi_i(e) + \varepsilon \right) \\
\delta_{\text{addition}} &:= \left( \mathcal{S} \mapsto \mathcal{S} \cup \{ e_{\text{new}} \} \right) \\
\delta_{\text{removal}} &:= \left( \mathcal{S} \mapsto \mathcal{S} \setminus \{ e \} \right) \\
\delta_{\text{replacement}} &:= \left( \mathcal{S} \mapsto (\mathcal{S} \setminus \{ e \}) \cup \{ e' \} \right)
\end{align}

All operations must be performable while preserving phenomenological consequences, such as the adaptation of occlusion, since even minimal intolerable changes in data quality can compromise the trustworthiness of statements derived from data with synthetic modification. The scientific community faces multiple challenges in acquisition or synthesizing LiDAR PCDs that are both realistic as well as reflect this compositional variance.

\section{State of the Art and Related Work}
\label{sec:sota}

\subsection{LiDAR Perception Applications}
Unlike RGB camera systems, which estimate the depth of environments, LiDAR sensors directly measure the three-dimensional structure and provide spatial as well as material-specific information. The KITTI dataset \cite{kitti} enabled LiDAR-based perception development using the Velodyne HDL-64E with 2.2 million points per second. As modern sensors like the Ouster OS1-128 Rev. 7 achieve up to 5.2 million points per second, and no public dataset matches these properties, developers must record their own labeled datasets.
LiDAR perception has progressed from 2D BEV projections to voxelized, point-based, and spherical projection methods, which structure range, XYZ, and intensity into tensors for real-time inference. Core tasks include 3D object detection~\cite{pvrcnn}, semantic and instance segmentation~\cite{squeezeseg}. Recent methods fuse LiDAR with other sensors to improve accuracy and robustness ~\cite{liu2022bevfusion}.

\subsection{Challenges in Empirical Data Acquisition}
Acquiring real world data through on-road tests, as commonly provided in datasets such as KITTI~\cite{kitti} Dataset, fundamentally limits the ability to maintain \textit{ceteris-paribus} conditions: even minimal changes in one factor (e.g., the position of a vehicle) often induce additional, uncontrolled variations (e.g., appearance of objects, lighting conditions, weather). Hence, naturally occurring data is biased toward frequently encountered scenes, while rare but safety-critical situations remain underrepresented. To address these issues, various research groups employ \emph{parameter-driven} methodologies, where real world drives are either pre-planned or dynamically adapted based on specific criteria (e.g., traffic density, surface characteristics, local geometric properties). For instance, geometric descriptors like \textit{Surface Variation} can be solved analytically from a covariance matrix $\Sigma_i$ of a PCD, where $\lambda_1 \geq \lambda_2 \geq \lambda_3$ denote the eigenvalues of $\Sigma_i$. ~\cite{jan_2016_CVPR}:
\vspace{+0.2mm}
\begin{align}
\Sigma_i 
= \frac{1}{N} 
  \sum_{n \in \mathcal{P}^N} 
  \bigl(\mathbf{p}_n - \bar{\mathbf{p}}\bigr)
  \bigl(\mathbf{p}_n - \bar{\mathbf{p}}\bigr)^\top
\end{align}
\vspace{-3mm}
\begin{align}
\text{Surface Variation} 
= \frac{\lambda_3}{\lambda_1 + \lambda_2 + \lambda_3},
\end{align}
\vspace{-0.1mm}

These descriptors can serve as control variables for route selection to more specifically capture relevant environmental factors, for example through a geometry-aware sampling strategy ~\cite{Heide2020b}. Nevertheless, continuous real world data collection remains systematically biased towards frequently occurring conditions, such as the stochastic distribution of objects and their positions. For example, cases in which the surface variation of an object becomes similar to that of its immediate surroundings result -- under an information-theoretic perspective -- in unstructured point constellations, which inherently challenge classification algorithms.
Since operational parameters such as object attributes or spatial configurations cannot be directly controlled, enforcing \textit{ceteris paribus} conditions is practically infeasible for empirically acquired data. The number of required permutations of semantic factors increases combinatorially. As argued in~\cite{liu2022curserarityautonomousvehicles}, the use of simulation is therefore essential to compensate for this imbalance and to systematically expose models to low-probability but functionally critical situations.

\subsection{Challenges with Virtual LiDAR Models}
\label{sec:challenge_virtual}

\begin{figure} 
\centering \includegraphics[width=0.6\linewidth]{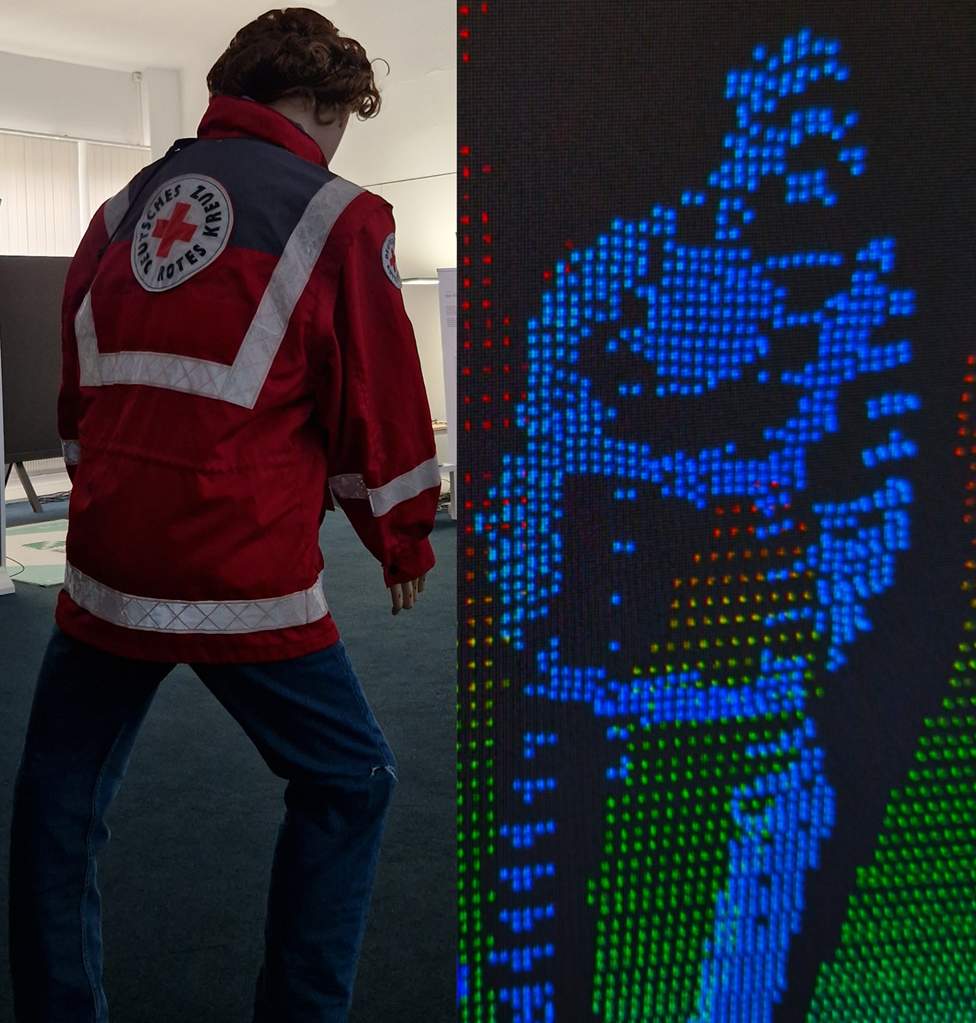} \caption{Reflectors on paramedic jacket causing object-exclusive local overexposure effects, canceling points on nearby less-reflective areas.} 
\label{fig:refl-phenomena} 
\end{figure}

Virtual 3D environments allow controlled variation and provide computed rays for model-based LiDAR simulation, but they simplify real-world scenes and sensor behavior, including ray generation~\cite{Vargas_Rivero_2024}. Virtual models replicate only analytically defined or empirically known effects. Accurate ray simulation requires detailed 3D environments, which are costly to design, even for single street segments. While automated generation~\cite{Huang_2024} reduces effort, it struggles with geometric accuracy over distance, an issue for LiDAR with ranges up to 200 m.
Approximating real sensor responses requires wavelength-specific reflectivity and absorption measures, obtained through reference measurements and integrated into the simulation \cite{Pharr_Jakob_Humphreys_2016, Raj_2020}. Integrating such measures is challenging due to partially non-linear correlations, including wavelength-dependent reflectivity, angle of incidence, surface roughness, and distance-based attenuation. High-fidelity intensity simulation requires high-resolution surface maps with per-pixel annotations of backscatter and micro-roughness. Physically-based reflectance models like BRDF and BSDF are used to describe direction-dependent scattering and transmission \cite{Pharr_Jakob_Humphreys_2016}. Real sensors are affected by intrinsic (e.g., photodiode, amplifier noise) and extrinsic noise (e.g., fog, rain, dust, sunlight). In simulations, such effects are typically approximated by stochastic models, which entail simplifications\cite{haider2023}. 
The central challenge in developing simulations is the simulation-to-reality gap: simulation results may appear accurate to humans but still deviate beyond acceptable thresholds, causing test object behavior to diverge as it would be confronted with real data. To build confidence that synthetic data are suitable for decision-making, virtual results must be validated against real sensor measurements. This validation should not be considered outside the overall development cost. Coupled or higher-order effects are particularly difficult to model due to the high dimensionality of influencing factors. Overconfidence in simulation completeness can obscure such effects, as some sensor-environment interactions remain untested under real-world conditions. In particular, underrepresented reference measurements can lead to \textit{unknown-unknowns} phenomena that have never been systematized due to a lack of prior measurements or theoretical explanation. This misplaced confidence may discourage further data collection, thereby hindering the minimization of blind spots in system understanding. Consequently, test objects approved solely through simulation may exhibit failures when exposed to actual operational environments. Additionally, if test objects behave worse in simulation, e.g., by producing false positives that do not occur with real sensor input, this can lead to a loss of trust in simulation-based development. 
Figure \ref{fig:refl-phenomena} illustrates a typical example of an \textit{unknown unknown}: highly reflective elements on a paramedic response jacket cause local overexposure in the real sensor return (right image), leading to point cancellation effects in neighboring regions with lower reflectivity. We compared several commercially available environmental simulations with a re-simulated, parametrized OS1 sensor model on different objects exhibiting reflective properties, including paramedics. In all tested simulations, only the intensity of reflective points was increased, but no point cancellation effects were observed. This indicates that current virtual simulations are still challenged in accurately reproducing reflectivity phenomena.
\subsection{Metrics for Comparing Point Cloud Data}

To evaluate the similarity between point clouds, such as real to simulated ones,
following metrics are established:\\

\vspace{-2mm}
The \textbf{Chamfer Distance} measures the average squared distance between points in two point clouds. Small values indicate good average geometric agreement.
\vspace{-1mm}
\begin{equation*}
CD(P, Q) = \\
\end{equation*}
\vspace{-7mm}
\begin{align}
    \frac{1}{|P|} \sum_{p \in P} \min_{q \in Q} \|p - q\|^2 + \frac{1}{|Q|} \sum_{q \in Q} \min_{p \in P} \|q - p\|^2
    % \end{equation*}
\end{align}

The \textbf{Hausdorff Distance} captures the maximum deviation between closest points in each set.
\begin{equation*}
HD(P, Q) = 
\end{equation*}
\vspace{-9mm}
\begin{align}
\max \left\{ \sup_{p \in P} \inf_{q \in Q} \|p - q\|, \sup_{q \in Q} \inf_{p \in P} \|q - p\| \right\}
\end{align}
\\
\textbf{Root Mean Square Error (RMSE)} quantifies average squared deviation based on nearest-neighbor distances. Suitable for general deviation evaluation in meters. Lower values indicate better fit.
\begin{align}
RMSE(P, Q) = \sqrt{\frac{1}{|P|} \sum_{p \in P} \min_{q \in Q} \|p - q\|^2}
\end{align}
\\
\vspace{-1mm}
To quantify the similarity between two point clouds $P$ and $Q$ the \textbf{F1-Score} is calculated. Based on a symmetric nearest-neighbor criterion with a threshold \( \tau > 0 \), precision and recall is defined as follows:
\begin{align}
\text{Precision} &= \frac{1}{|P|} \sum_{\mathbf{p} \in P} \mathbf{1} \left( \min_{\mathbf{q} \in Q} \| \mathbf{p} - \mathbf{q} \|_2 < \tau \right), \\
\text{Recall} &= \frac{1}{|Q|} \sum_{\mathbf{q} \in Q} \mathbf{1} \left( \min_{\mathbf{p} \in P} \| \mathbf{q} - \mathbf{p} \|_2 < \tau \right).
\end{align}
\( \mathbf{1}(\cdot) \) denotes the indicator function, which evaluates to 1 if the condition inside is true, and 0 otherwise.
The F1 score is then given by the harmonic mean of precision and recall:
\begin{equation}
\text{F1} = 
2 \cdot \frac{\text{Precision} \cdot \text{Recall}}{\text{Precision} + \text{Recall}}
\end{equation}
This metric captures both the point-wise completeness and accuracy of the predicted point cloud with respect to the reference, and is particularly well-suited for evaluating geometric consistency in 3D reconstruction and registration tasks.

\subsection{Target Objects for Perception Validation}
Target objects are commonly used in proving grounds to evaluate e.g., braking on obstacles in automated driving systems. Standardized dummies with defined reflectivity are applied in testing automatic emergency braking systems \cite{haran2016infrared}, and to compare LiDAR performance on human-like shapes \cite{lambert2020performance}. Dynamic tests employ systems like the 6D Target Mover for reproducible target motion \cite{messring_6d_target_mover}. In radar testing, fixed humanoid targets with varying clothing and poses are placed on or inside vehicles to analyze sensor responses. The results indicate that a realistic appearance, including shape and material properties, becomes increasingly relevant for material-sensitive sensors, such as radar ~\cite{Abadpour2023_1000160786} and consequently LiDAR.

\subsection{Augmentation for Dataset Diversification}
Object detection performance depends not only on object attributes but also on the surrounding scene context. To diversify real world LiDAR data for both training and testing, two main augmentation strategies are commonly used. Both insert object point clouds into real scenes and require plausible removal of points that would be occluded by the inserted object. The first group of methods, such as LiDAR-Aug \cite{Fang_2021_CVPR}, uses mesh--based ray tracing and virtual sensors to synthesize new points. Occlusion is modeled geometrically, but the simulated data is limited by the fidelity of the sensor model and environmental assumptions.
The second group, including Real3D-Aug \cite{sebek2022real3daug}, applies copy-paste augmentation with real object point clouds from within the dataset. Paved2Paradise (P2P) \cite{paved2paradise} adopts the copy-paste principle of Real3D-Aug and applies it to outdoor scenes, using separately recorded real world objects and backgrounds. While it takes a first step toward controlled augmentation, it remains susceptible to uncontrolled factors such as sunlight or environmental noise and requires clarification on scalability. Point-wise occlusion computations, estimated either morphologically or by inverse approaches, can be effective but lack a direct physical modeling, particularly in sparse, noisy, or overexposed data, as exemplified in section \ref{sec:challenge_virtual}. Point-based occlusion analysis becomes invalid if the point clouds contain significant noise or artifacts, as missing point clusters lead to inconsistent occlusion modeling, thereby contributing to a simulation-to-reality gap. The use of full 3D meshes, such as those employed in virtual augmentation, could remedy this, as occlusion modeling through geometric raycasting, can be independent of noise or artifacts of a point cloud. However, this requires that the mesh is correctly aligned to the object's position in the point cloud to ensure spatial consistency.

\section{Concept}
\label{sec:concept}
Current testing strategies show that existing LiDAR perception tests rely mainly on either real-world driving or virtual simulations. A potential complement lies in bridging these domains: Hardware simulations remain underrepresented in form of proving grounds in this context, which can feasibly only vary standardized objects, limiting the range of testable cases. The recent augmentation methods may serve as an enabler by allowing physically measured target objects, such as those similar to proving grounds, to be seamlessly and plausibly integrated into required real scenes. \textit{Point Cloud Recombination (PCR)} builds on the idea of real data augmentation in terms of generating test data for validation purposes. Analogous to genetic recombination, we systematically dissect elements from different sources in laboratory and recombine them into plausible variants. PCR modifies empirically acquired scenes at scale by inserting real data from customizable target objects captured in laboratory conditions, while maintaining plausibility with respect to the constraints of a specific sensor under test. In addition, while training data may include unrealistic samples to support generalization \cite{tremblay2018trainingdeepnetworkssynthetic}, validation requires data that remain within the plausible output range of real sensors to ensure result legitimacy. This leads to three key requirements:

\begin{itemize}
    \item (Req 1) The sensor under test must be mounted in its designated installation position as defined for later deployment, ensuring that the data reflect realistic system conditions.
    \item (Req 2) Object point clusters must be inserted within the measured radial distance to the sensor and along discrete orbital path, maintaining their original orientation to preserve geometric plausibility.
    \item (Req 3) If specific conditions from a real scene (e.g., overexposure) cannot be recreated, the associated object data are incompatible for insertion.
\end{itemize}
PCR acknowledges these constraints and promotes the idea that measurements should be conducted in laboratory environments, where external conditions can either be excluded or deliberately and reproducibly introduced. Moreover, current approaches to computing occlusion encounter significant limitations when handling artifacts present in LiDAR scans. To mitigate this, we use high-fidelity reconstructed 3D meshes from images registered to object point clouds to preserve occlusion behavior, even when real points are missing or distorted. Since validation requires datasets tailored to the sensor under test, scalable acquisition solutions are essential. We therefore developed an automated, modular, and dimensionally scalable laboratory setup. This hardware--simulative environment is complemented by an expert--oriented automation pipeline, enabling efficient and reproducible test data generation under controlled conditions. Figure \ref{fig:pcr_intro} illustrates the method in following steps:\\

\textbf{PCR 1. Empirical Data Acquisition:}\\
We recommend beginning with empirical data collection e.g., driving in the target domain, as it simplifies subsequent laboratory calibration. To detect the compatibility between other measures relevant for the ODD, data such as vibration, humidity, and lighting exposure are continuously recorded. These data are important because they may aid future data collection improvements, posterior data filtering and debugging purposes. Finally, the recording of PCD Scenes for gaining valid-able results demands that the sensor under test must be mounted in its designated installation position within the target system. The whole PCD scenes from the sensor are recorded and stored in a database and captured scenes are cataloged with meta descriptions.

\textbf{PCR 2. Systematic Data Acquisition in Laboratory:}\\
Decoupled from real world driving, the setup ensures controlled conditions and allows targeted variation of object characteristics. The LiDAR sensor is typically mounted at vehicle-equivalent height to preserve realistic acquisition geometry, while the target objects are moved to their target position manually. Alternatively, depending on the use case, the sensor may be mounted on a mobile robotic platform, while the target object remain stationary. This variant depends on the relevancy, target object and measurement area size and automates manual placing of target objects: The mobile platform is controlled by a state machine adjusting spatial parameters such as distance, position, and yaw to enable automated positioning before sample capturing. \\
\textit{PCR 2.a) 3D-Reconstruction via 360-degree Images:} For occlusion modeling, each object is captured through 360-degree drone imagery and reconstructed into a solid 3D mesh using techniques such as photogrammetry, neural radiance field\cite{mildenhall2020nerfrepresentingscenesneural} with subsequent surface computation. Drones can be automated for rotational movements and typically include gimbals for image stabilization, facilitating the efficient generation of meshes with acceptable quality. To ensure consistency with the acquired point cloud data, each target object type--meaning every change in shape or material--requires a corresponding mesh.\\
\textit{PCR-2.b) Capture Object with Real LiDAR:} The laboratory environment offers advantages compared to augmenting data from explorative real world data, where objects must be manually segmented. In the laboratory setup, the target object is placed in an open area, allowing the acquisition of object-related point cloud data directly through a single predefined 3D box shaped region of interest, where a target object can occur. In parallel, metadata is generated automatically, including the relative X, Y, and Z distance from the sensor center to the object center, as well as size and orientation information for object distinction and automated annotation.

% ---------- ALGORITHM 1 ----------
\begin{algorithm}[H]
\caption{Mesh-to-Point-Cloud Registration through multiple iterative closest point (ICP)}
\label{alg:icp_registration}
\begin{algorithmic}[1]
\Require Mesh $\mathcal{M}$, point cloud $\mathcal{P}^{xyz}_{\text{obj}}$, divisor $d$
\Ensure Aligned mesh $\mathcal{M}^*$
\State $D^* \gets \infty$, $\mathcal{M}^* \gets \emptyset$
\For{$\theta \in \{0, \frac{360}{d}, \dots, 360{-}\frac{360}{d}\}$}
  \State Rotate a copy $\mathcal{M}_\theta$ of $\mathcal{M}$ around $\theta$ (Z-axis)
  \State Sample $30{,}000$ points $\mathcal{M}_\theta^{pc}$ 
  \State estimate $T_0$ from centroids
  \State $T \gets \text{ICP}(\mathcal{M}_\theta^{pc}, \mathcal{P}^{xyz}_{\text{obj}}, T_0, [0.1, 0.05])$
  \State Apply $T$ to $\mathcal{M}_\theta$, compute $D_\theta$
  \If{$D_\theta < D^*$} \State $D^* \gets D_\theta$, $\mathcal{M}^* \gets \mathcal{M}_\theta$ \EndIf
\EndFor
\State \Return $\mathcal{M}^*$
\end{algorithmic}
\end{algorithm}
% ---------- ALGORITHM 2 ----------
\begin{algorithm}[H]
\caption{Occlusion Computation and Point Removal}
\label{alg:shadowcasting}
\begin{algorithmic}[1]
  \Require point cloud $\mathcal{P}_{\text{scene}}^{xyz} \subset \mathbb{R}^3$, registered $\mathcal{M}^*$
  \Ensure Occluded point coordinates $\mathcal{P}_{\text{occ}}^{xyz}$
  \State $\mathbf{O} \gets \mathbf{0} \in \mathbb{R}^{|\mathcal{P}_{\text{scene}}^{xyz}| \times 3}$ \Comment{Ray origins}
  \State $\mathbf{R} \gets [\mathbf{O}, \mathcal{P}_{\text{scene}}^{xyz}] \in \mathbb{R}^{|\mathcal{P}_{\text{scene}}^{xyz}| \times 6}$ \Comment{Concatenate rays}
  \State Create raycasting scene and insert mesh $\mathcal{M}$
  \State Cast rays $\mathbf{R}$ and store hit distances $t_{\text{hit}}$
  \State $I_{\text{hit}} \gets \{\,i \mid t_{\text{hit}}[i] < \infty \,\}$
  \State $\mathcal{P}_{\text{occ}}^{xyz} \gets \{\, \mathcal{P}_{\text{scene}}^{xyz}[i] \mid i \in I_{\text{hit}} \,\}$
  \State \Return $\mathcal{P}_{\text{occ}}^{xyz}$
\end{algorithmic}
\end{algorithm}
% ---------- ALGORITHM 3 ----------
\begin{algorithm}[H]
\caption{Adding Object PCD After Occlusion Removal On Scene PCD}
\label{alg:pcd_fusion}
\begin{algorithmic}[1]
  \Require Full scene PCD $\mathcal{P}_{\text{scene}}$ (with attributes), object PCD $\mathcal{P}_{\text{obj}}$, occluded XYZ points $\mathcal{P}_{\text{occ}}^{xyz}$
  \Ensure Recombined PCD $\mathcal{P}_{\text{recomb}}$
  \State Filter scene by removing points matching $\mathcal{P}_{\text{occ}}^{xyz}$ in XYZ dimensions
  \State $\mathcal{P}_{\text{filtered}} \gets \mathcal{P}_{\text{scene}} \setminus \mathcal{P}_{\text{occ}}^{xyz}$
  \State Concatenate with object PCD: $\mathcal{P}_{\text{recomb}} \gets \mathcal{P}_{\text{filtered}} \cup \mathcal{P}_{\text{obj}}$
  \State \Return $\mathcal{P}_{\text{recomb}}$
\end{algorithmic}
\end{algorithm}

\textbf{PCR 3. Raw Data Processing:}
Following the acquisition of PCDs from both real world scenes and laboratory objects, automated processing pipelines can be initialized based on object-specific properties such as type, position, orientation, and pose. Experts can define multiple target configurations in which concrete object instances are inserted into selected scenes. To ensure plausibility, a compatibility analysis is performed using an insertion map that encodes valid placement regions. This map verifies that objects do not unrealistically collide with the scene and can be extended with further constraints to fulfill specific test case requirements. As long as object to scene combinations comply with the defined placement rules, multiple objects can be sequentially inserted into one or more scenes.

\begin{figure*}
    \centering
    \includegraphics[width=0.75\linewidth]{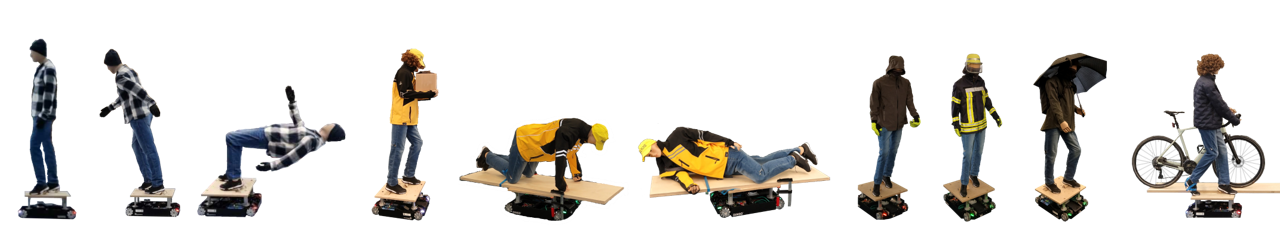}
    \vspace{-5pt}
    \caption{Collection of humanoid target objects used for application and evaluation. From left to right: 
    \textit{casual\_clothed\_pedestrian\_series} (standing, running, falling), 
    \textit{postman\_series} (carrying\_parcel, crouching, lying),
    \textit{pedestrian\_with\_reflective\_gloves}, 
    \textit{firefighter}, 
    \textit{person\_with\_umbrella\_and\_bag}, 
    \textit{person\_pushing\_bicycle}.}
    \label{fig:targets}
\end{figure*}
\textbf{PCR 4. Single Point Cloud Recombination:}\\
For each frame, the recombination processes a scene, a PCD and a corresponding mesh through the following steps to generate a recombined point cloud:

\textit{PCR 4.a) Mesh-to-Point-Cloud Registration:} Algorithm \ref{alg:icp_registration} prepares an object-type-specific mesh for occlusion-aware integration by spatially aligning it to an empirically acquired point cloud of the same object type. Since the mesh is defined in its own local coordinate system, it must be transformed into the spatial configuration that matches the future placement of the point cloud within a scene. To achieve this, we perform multiple ICP registrations with rotational sampling along the Z-axis, using the Chamfer Distance to evaluate the alignment quality of each attempt. The mesh is scaled to match the height of the point cloud and refined through a two-stage ICP process with progressively stricter thresholds.
\\
\textit{PCR 4.b) Occlusion Computation and Point Removal:} The spatially 3D-transformed mesh is inserted into a virtual raycasting environment, together with the point cloud scene, both aligned in a common sensor-centered coordinate system (see Algorithm \ref{alg:shadowcasting}). To compute occlusions, rays are cast from each scene point towards the sensor origin. Points whose rays intersect the mesh are marked as occluded and subsequently removed by indexing their coordinates in the original point cloud. All additional point attributes (e.g., intensity) remain preserved.
\\
\textit{PCR-4.C) Integration of PCD Object:} The object PCD is inserted into the occlusion-filtered scene afterwards (see Algorithm \ref{alg:pcd_fusion}). Because the geometry, pose, and placement are inherited from the measurement setup, no further estimation is necessary. Meta data including insertion parameters (e.g., position, orientation, mesh ID) can be used for automated label generation, such as 3D Bounding Boxes or comprehensive object lists.

\section{Application}
\label{sec:application}

\begin{figure}
    \centering
    \includegraphics[width=0.75\linewidth]{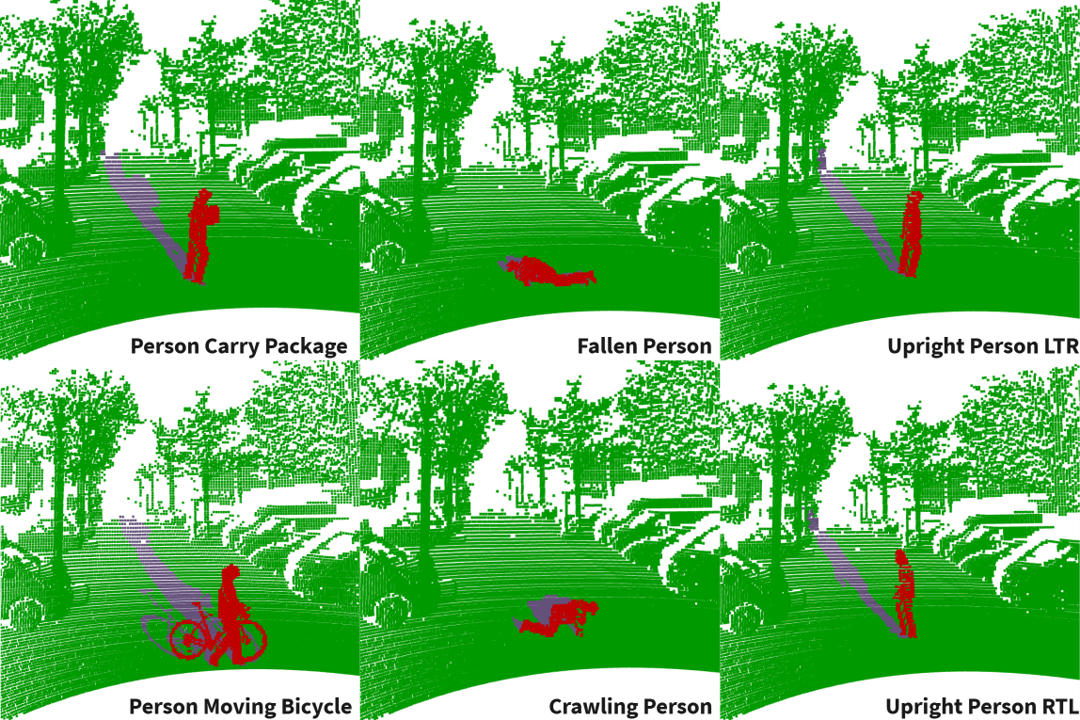}
    \vspace{-5pt}
    \caption{Applying PCR for controlled variation of object attributes (e.g., pose, props, hairstyle) while keeping the scene constant.}
    \label{fig:vary_pose}
\end{figure}

\begin{figure}
    \centering
    \includegraphics[width=0.75\linewidth]{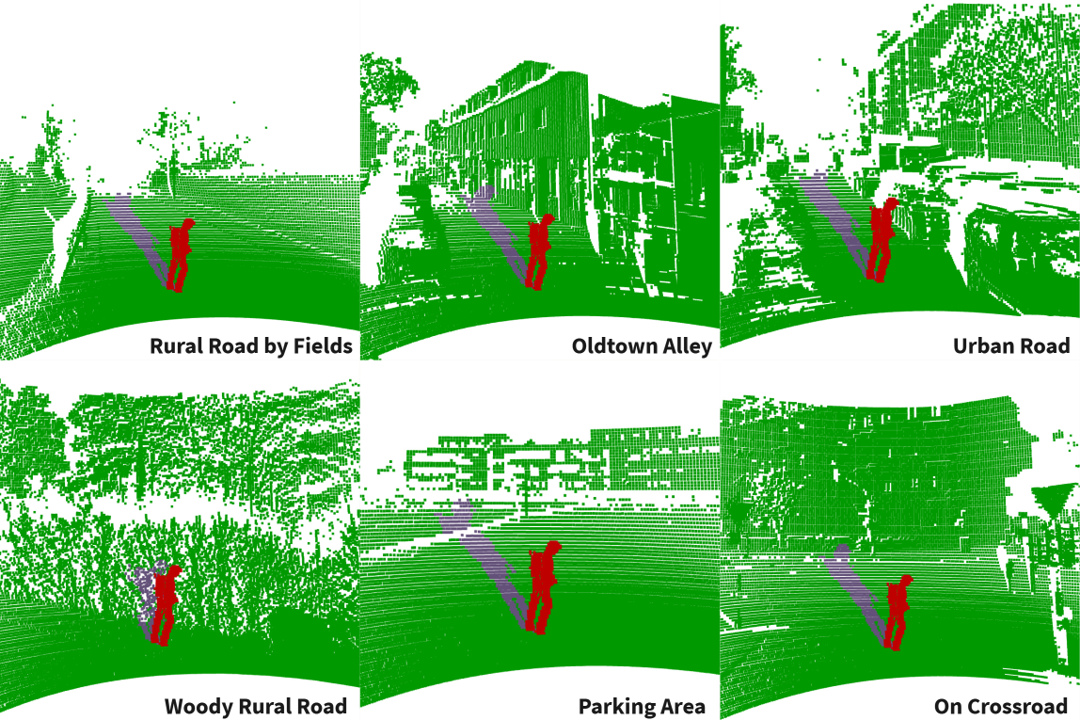}
    \vspace{-5pt}
    \caption{Applying PCR for controlled variation of the scene background while preserving the object's pose and attributes}
    \label{fig:vary_scene}
\end{figure}

For this study, we validate a LiDAR-based reference system using an Ouster OS1-128 Rev.~7 sensor mounted at 1.80 m on a HAD research vehicle. The focus lies on evaluating the robustness of pedestrian detection in public traffic environments.
To reflect a broad spectrum of European conditions, outdoor LiDAR data was recorded in various road types. These include modern urban areas, suburban and village roads, historical city centers with narrow alleys and irregular facades, and rural areas with dense vegetation and high surface variation. For this application of PCR, our insertion map computation is realized by computing insertion areas on planar regions using eigenvalue-based surface variation analysis with adaptive radius settings (adapted from CloudCompare).
Object-specific data acquisition was conducted in controlled indoor environments: a laboratory hall (up to 10,0 m range between sensor center to object center) and a sports hall (up to 35 m). An omnidirectional mobile robot (Clearpath Dingo) was used to precisely position dressed mannequins synthesizing pedestrians. As realistic replication of human skin reflectivity was not possible, mannequins were therefore fully clothed. Furthermore, props such as bags and umbrellas were used to simulate realistic urban scenarios. The robot localization employed SLAM (GMapping and AMCL) using a secondary LiDAR (Ouster OS0-64). A ROS-based state machine controlled navigation, and each Object PCD was captured with an ROI and annotated with 3D centroid-based relative translation. High-fidelity 3D meshes of mannequin configurations were reconstructed using 360-degree drone footage (DJI Mini 4) processed via LumaAI. All computations were performed on an NVIDIA RTX 4090 GPU, with each point cloud integration requiring approximately 1.8 s.
Figure \ref{fig:targets} shows the prepared target object types. A humanoid mannequin in casual clothing was configured in three poses: standing, walking, and falling, with distances ranging from 4 up to 35 meters to benchmark the Mesh-to-PointCloud registration. A second series represents a postal worker in three synthesized poses: carrying a parcel, crouching, and lying down. These setups shall demonstrate that even rare, safety--critical scenarios can be realistically reproduced under controlled conditions using real sensor data. To increase structural variability, additional object constellations were introduced, including a person holding an umbrella and one pushing a bicycle--configurations often underrepresented in public datasets. To explore the impact of surface reflectivity, we included a mannequin with reflective markers on the hands and a firefighter figure with approximately 40 \% reflective surface area on the upper body. As illustrated in Figure~\ref{fig:vary_scene}, PCR enables \textit{ceteris-paribus} testing by keeping the object fixed while varying the background. Conversely, Figure~\ref{fig:vary_pose} holds the scene constant while changing the object pose or equipment. PCR shows its capability to generate trustful data which enable determination whether failures in perception are caused by an attribute of an object, its surrounding scene, or their combination.

\section{Evaluation}
\label{sec:eval}
\begin{figure}[H]
    \centering
    \includegraphics[width=0.75\linewidth]{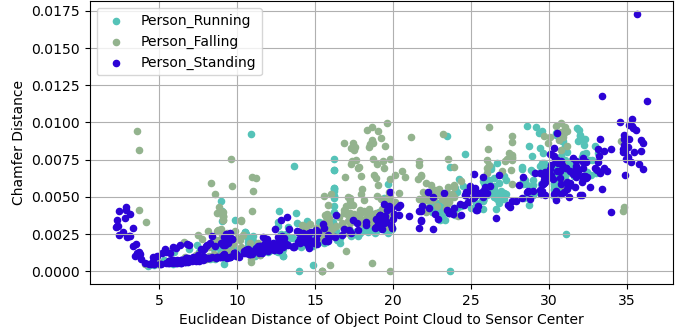}
    \vspace{-4 mm}
    \caption{Chamfer Distance between reconstructed 3D meshes and LiDAR point clouds as a function of the euclidean distance to the sensor center, evaluated for the casual\_clothed\_pedestrian series}
    \vspace{-1 mm}
    \label{fig:icp_acc}
\end{figure}

Firstly, the evaluation focuses only on the ICP-based object-to-mesh registration. Therefore, the casual\_clothed\_pedestrian series were evaluated across three poses (standing, running, and falling) at distances up to 35 m. The registration accuracy was measured using the best Chamfer Distance which is already part of the algorithm. The alignment accuracy and precision between reconstructed meshes and LiDAR point clouds correlated to euclidean distance is depicted in Figure \ref{fig:icp_acc}. The standing pose showed the highest accuracy (mean: 0.0031) and precision (std: 0.0012). In contrast, the running (mean: 0.0048, std: 0.0021) and falling poses (mean: 0.0045, std: 0.0020) yielded lower consistency, indicating pose-related variability. The Chamfer Distance increased with the relative distance to the sensor across all poses, reflecting reduced point density and registration reliability. This trend was least pronounced in the standing pose. As Chamfer Distance does not penalize mirrored alignments, especially in symmetric shapes, we included manual verification. Rotational sampling around the vertical axis combined with Chamfer-based selection reliably avoided such misalignment in all cases.

\vspace{-4 mm}

\begin{figure}[H]
    \centering
    \includegraphics[width=0.75\linewidth]{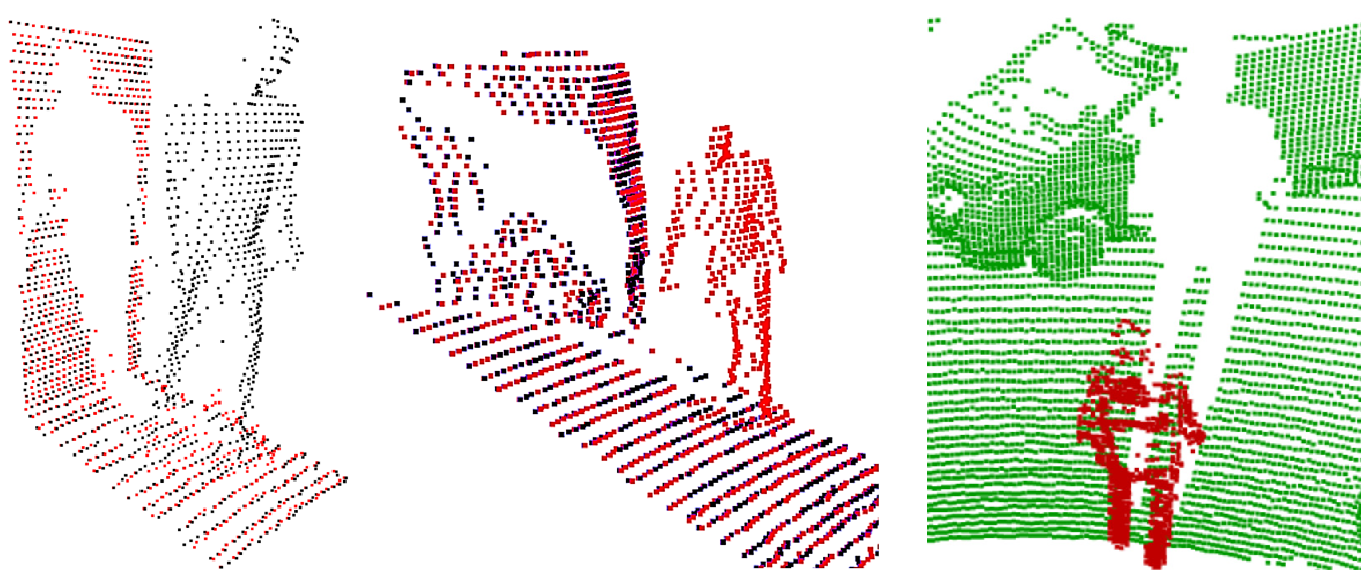}
    \caption{Pruned results of PCR: Highlighting the similarity between the reference point cloud (black) and the recombined point cloud (red) of a target object placed in front of a wall (left) and a car (middle). The right image shows a humanoid (red) with reflective clothing that have missing points; PCR nonetheless enables plausible occlusion handling on the scene (green).}
    \label{fig:roi_pruned_pcds}
\end{figure}

\begin{table}[H]
\centering
\begin{tabular}{lcccc}
\toprule
 \textbf{Background} & \textbf{Chamfer} & \textbf{Hausdorff} & \textbf{RMSE} & \textbf{F1-Score}  \\
\midrule
\textbf{Wall (SV 0.11)} & 0.0006 & 0.1588  & 0.0166 & 99.7027 \\
\textbf{Car (SV 0.28)} & 0.0011 & 0.2498 & 0.0264 & 99.1920 \\
\textbf{Bicycles (SV 0.29)} & 0.0032 & 0.8647  & 0.0394  & 99.2862 \\
\textbf{Tree (SV 0.33)} & 0.0161 & 0.0863  & 0.0242 & 99.1124 \\
\textbf{Bushes (SV 0.34)} &  0.0019 & 0.6048  & 0.0268 & 98.8108 \\
\midrule
\textbf{Mean All} & 0.0045 & 0.39288  & 0.02668 & 99.02082 \\
\textbf{Noise Reference} & 0.0003 & 0.12303  & 0.0125 & 99.9602 \\
\bottomrule
\end{tabular}
\vspace{1 mm}
\caption{Quantification of dissimilarities for selective pruned reference to recombinated scenes, sorted by their averaged Surface Variation (SV)}
\vspace{-5 mm}
\label{tab:dissimilarities}
\end{table}
Secondly, the evaluation addresses the quantification of the simulation-to-reality gap in PCR results. To ensure valid mesh-based occlusion handling, target objects are not scanned in isolation but physically placed at their intended position within the scene. A reference scan is taken with the object present, followed by a second, object-free scan used as input for the recombination process. The object region is replaced by inserting the measured object points from the reference and applying occlusion modeling based on a drone-derived mesh. Both resulting scenes are pruned to retain only the object and its surrounding occlusion volume (see Figure~\ref{fig:roi_pruned_pcds}). To establish a baseline for assessing dissimilarities introduced by PCR, we define a Noise Reference, based on two LiDAR scans taken seconds apart under identical conditions. Deviations within this noise range are considered indistinguishable from those of real scans and therefore do not compromise the plausibility of the recombined data. Table~\ref{tab:dissimilarities} quantifies geometric dissimilarities between reference and recombined scenes, sorted by the mean Gaussian surface variation (SV) of the background. Chamfer distances range from 0.0006 (Wall, SV 0.11) to 0.0161 (Tree, SV 0.33), with the Noise Reference baseline at 0.0003. RMSE values remain similarly close. Hausdorff distances are environments with higher SV (e.g., 0.8647 for bicycles), consistent with localized structural differences. F1-scores stay high across all backgrounds (98.81\% - 99.70\%), with a slight decrease observed as surface variation increases. The consistently high similarity metrics, complemented by visual inspection showing human perceptual indistinguishability of object contours, indicate that the recombined data deviates only minimally and is suitable for use in validation contexts.

\section{Conclusion}
\label{sec:conclusion}
Point Cloud Recombination (PCR) enables the targeted modification of real LiDAR scenes by integrating physically measured point clouds from target objects. Automated acquisition in laboratory environments allows controlled and scalable test data acquisition over a variety of objects and their properties. By registering a corresponding 3D mesh during recombination, the method supports phenomenon-aware occlusion modeling, in the presence of artifacts or sparse points on object PCD. Evaluation shows that PCR results exhibit only minimal deviations from real reference data, confirming suitability of the method  for systematic and validation-oriented test data generation.
The use of directly usable, measured LiDAR point clouds avoids the computationally intensive generation of virtual point clouds and the associated process effort for their qualification with regard to the simulation-to-reality gap. Given the necessity of both real data collections and hardware-related tests, PCR facilitates the generation of additional knowledge from these activities. The application of PCR has been demonstrated to assist in the reduction of uncertainties associated with physical sensor phenomena and hardware integration issues. This is becoming increasingly important, particularly for future applications such as logistics robots or even humanoid robots with close human-machine interaction.
Limitations remain for transparent surfaces and multi-path reflections between objects and the environment. Moreover, in the current setup, recombination was only feasible under moderate environmental conditions, without strong sunlight, precipitation, or large temperature differences, highlighting the need for test benches capable of simulating such conditions, e.g., through artificial sunlight systems. Future work includes evaluating PCR under environmental conditions involving significant vibrations, in combination with vibration test benches. Another direction involves multi-modal validation with physical target objects placed in front of LED wall systems, as used in modern virtual production environments, to enable synchronized camera and LiDAR hardware simulation. Given the absence of publicly available reference datasets for the Ouster OS1 sensor, PCR will also be extended into an automated pipeline to generate large-scale training and test data, for instance for 3D object detection tasks. Overall, PCR avoids the binary choice between fully virtual and fully real testing environments. Instead, it adopts a hybrid approach, deliberately dancing between virtuality and reality to enable trustful, controllable data generation. This facilitates failure cause analysis, supports the interpretability of perception systems, and advances their safety assessment.

\section{Acknowledgment}
The authors would like to specially thank the Institute of Sports and Sports Science (IfSS) at KIT, especially Mrs.~Beate Sewerin for allowing the measurements to take place, Mr.~Bishal Gautam for his valuable preparatory work and discussions. This work was supported by the German Federal Ministry for Economic Affairs and Climate Action within the project RepliCar with grant number 19A23002I.

\bibliographystyle{IEEEtran}
\bibliography{main}

\end{document}